\documentclass{article}

\usepackage{PRIMEarxiv}

\usepackage[utf8]{inputenc} % allow utf-8 input
\usepackage[T1]{fontenc}    % use 8-bit T1 fonts
\usepackage[hidelinks,allcolors=black]{hyperref}       % hyperlinks
\usepackage{url}            % simple URL typesetting
\usepackage{booktabs}       % professional-quality tables
\usepackage{amsfonts}       % blackboard math symbols
\usepackage{nicefrac}       % compact symbols for 1/2, etc.
\usepackage{microtype}      % microtypography
\usepackage{lipsum}
\usepackage{fancyhdr}       % header
\usepackage{graphicx}       % graphics
\newcommand{\regal}{{\textsf{REGAL}}}
\newcommand{\shc}{{\textsf{SHC}}}
\newcommand{\sa}{{\textsf{SA}}}
\newcommand{\ts}{{\textsf{TS}}}

\newcommand{\spatial}{{\textsf{SPATIAL}}}

\def\lcps{\textsf{X}}
\def\fcps{\textsf{Y}}

\def\graph{{\mathcal{G}}}
\def\vertexset{{\mathcal{V}}}
\def\edgeset{{\mathcal{E}}}

\def\assignment{{\mathbf{X}}}

\def\u{{\vphantom g}_}

\newcommand{\flip}{{\textsf{Flip}}}
\def\partition{\mathbf{\Pi}}

\newcommand{\rewrite}[1]{\textcolor{black}{#1}}
\def\stateR{{\mathbf{R}}}
\def\stateS{{\mathbf{S}}}
\def\constraint{{\mathbb{C}}}

\def\modelA{{\textsf{BAA}}}
\def\modelB{{\textsf{BCAA}}}
\def\modelC{{\textsf{AIO}}}

\usepackage{multirow}
\usepackage{mathtools}
\usepackage{diagbox}
\usepackage{nicefrac}
\usepackage{subcaption}
\usepackage{algpseudocode}
\usepackage[boxed, ruled, vlined]{algorithm2e}
\usepackage[noabbrev,capitalise]{cleveref}
\usepackage{comment}
\usepackage{xcolor}
\usepackage{commath}
\usepackage{latexsym}
\usepackage{booktabs}
\usepackage{microtype}

\title{Sampling-based techniques for designing school boundaries
\thanks{\textit{This work appeared as a part of doctoral dissertation } \cite{biswas2022spatial}.
} 
}

\author{
  Subhodip Biswas, Fanglan Chen \\
  Computer Science \\
  Virginia Tech \\
  Greater Washington, D.C., area\\
  \texttt{\{subhodip,fanglanc\}@vt.edu} \\
   \And
  Zhiqian Chen \\
  Computer Science \& Engineering  \\
  Mississippi State University \\
  Starkville, Mississippi\\
  \texttt{zchen@cse.msstate.edu} \\
  \AND
  Chang-Tien Lu, Naren Ramakrishnan \\
  Computer Science \\
  Virginia Tech \\
  Greater Washington, D.C., area\\
  \texttt{\{ctlu, naren\}@cs.vt.edu} \\
}

\begin{document}
\maketitle

\begin{abstract}
Recently, an increasing number of researchers, especially in the realm of political redistricting, have proposed sampling-based techniques to generate a subset of plans from the vast space of districting plans. These techniques have been increasingly adopted by U.S. courts of law and independent commissions as a tool for identifying partisan gerrymanders. Motivated by these recent developments, we develop a set of similar sampling techniques for designing school boundaries based on the flip proposal. Note that the flip proposal here refers to the change in the districting plan by a single assignment. These sampling-based techniques serve a dual purpose. They can be used as a baseline for comparing redistricting algorithms based on local search. Additionally, these techniques can help to infer the problem characteristics that may be further used for developing efficient redistricting methods. We empirically touch on both these aspects in regards to the problem of school redistricting. 
\end{abstract}

% keywords can be removed

The sampling-based techniques are used to compare the a given districting plan in context of a \textit{representative sample}, i.e., a set of valid alternative plans. Closely following this is the need to relate the sampling distribution to the criteria set forth by domain experts. This may be a tough ask since any redistricting effort can be accompanied by a varying set of criteria, some of which are difficulty to quantify objectively. As such our objective here is different. We would like to use a customized sampling distribution to generate an ensemble of plans and save the best quality plan, as determined by an objective function. In outlining the sampling-based approach, we closely follow the theory developed in \cite{recombination} and restate them here for comprehensiveness.

\begin{figure}
    \begin{center}
        \includegraphics[keepaspectratio, width=0.66\textwidth]{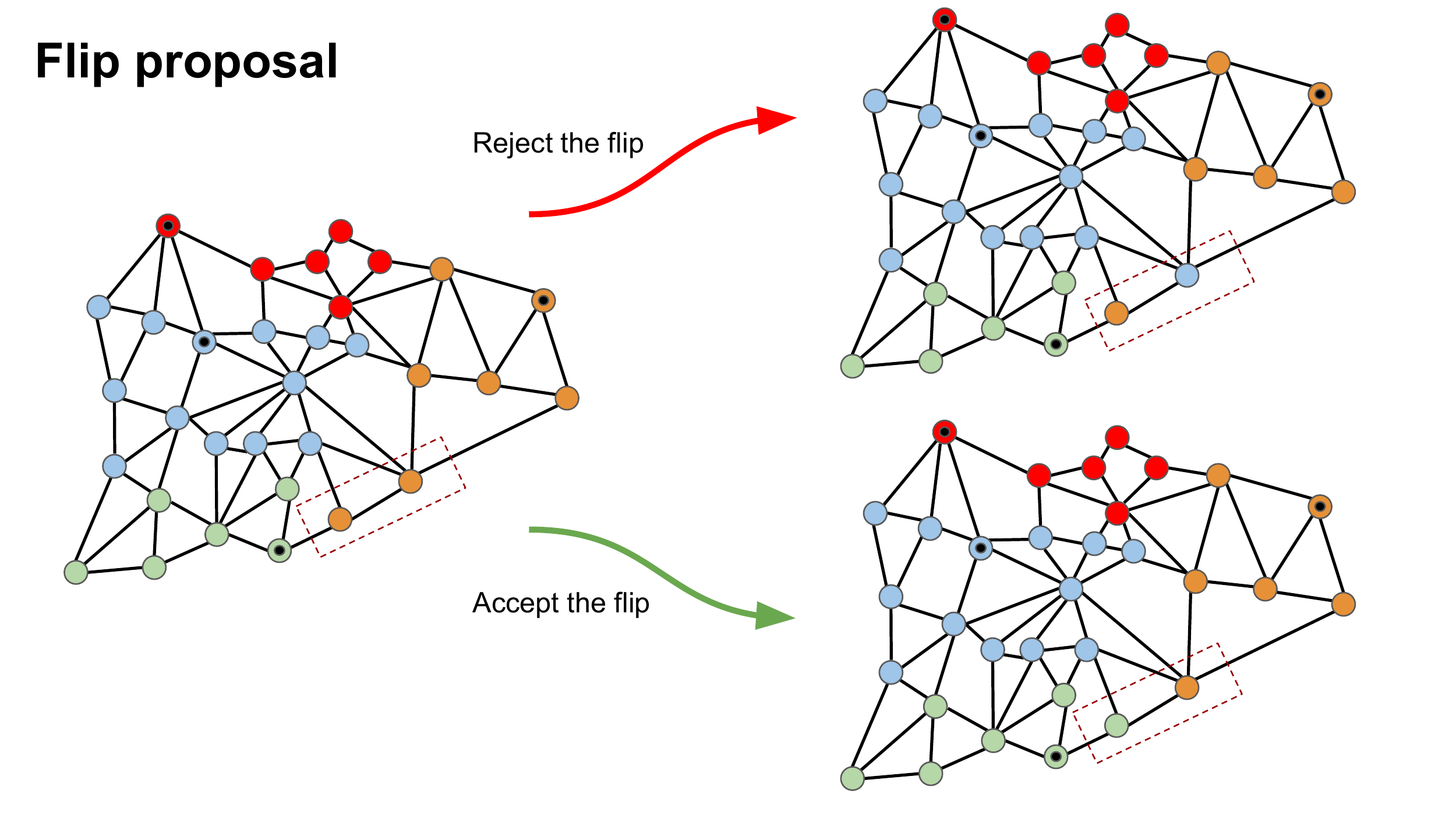}
    \caption{The flip proposal involves changing the assignment of a boundary node. }
    \label{ch4:fig:flip}
    \end{center}
\end{figure}

\begin{figure}
    \begin{center}
        \includegraphics[keepaspectratio, width=0.66\textwidth]{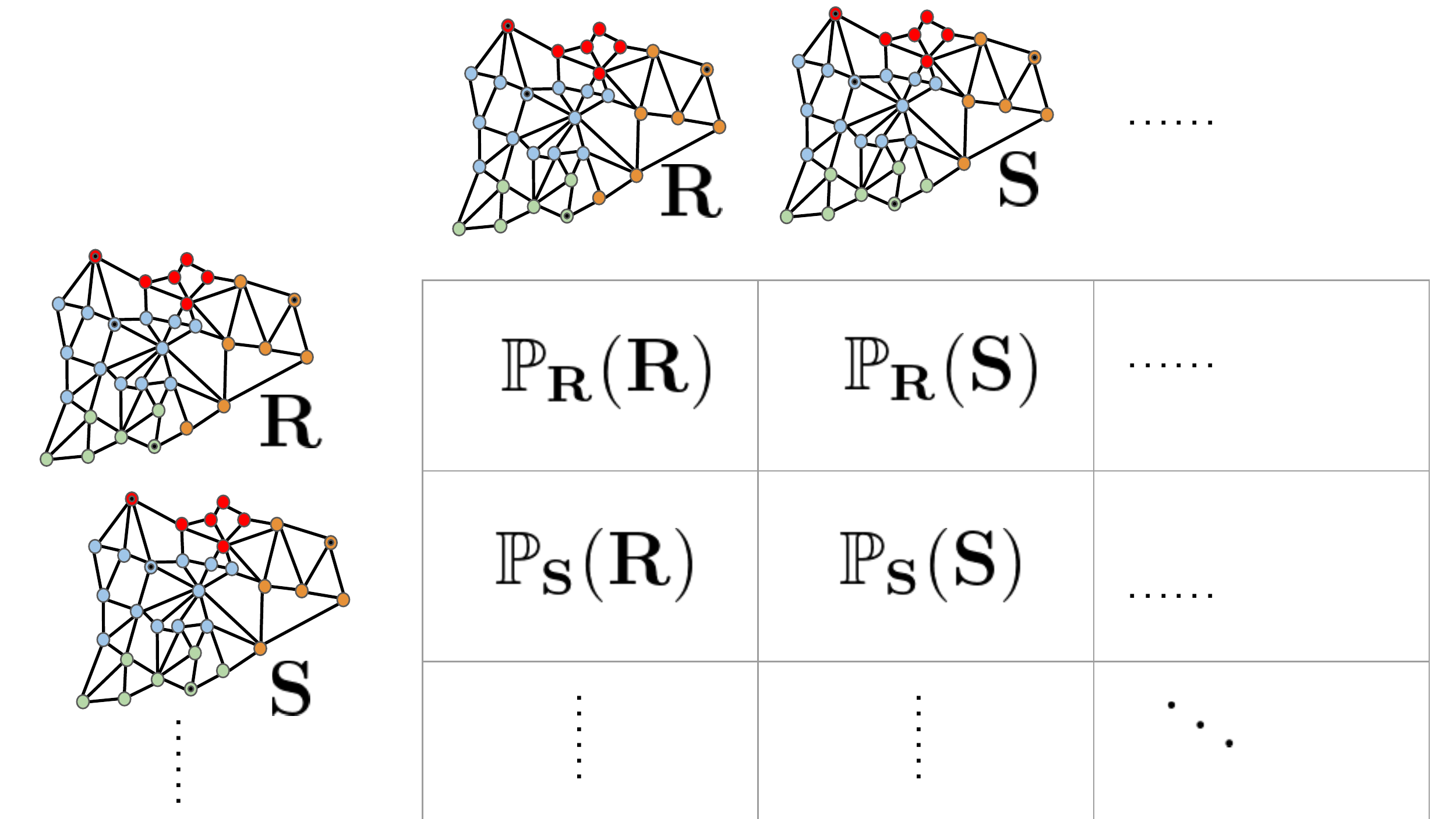}
    \caption{Theoretically if the flip proposal is carried out sufficient number of times, one may approximate the stationary distribution of transitions of the state space. Note that here each state correspond to a redistricting plan.}
    \label{ch4:fig:states}
    \end{center}
\end{figure}

Markov chain Monte Carlo, popularly known as MCMC, is an effective technique for sampling owing to strong underlying theory, in the form of mixing theorems and convergence properties~\cite{mcmc}. In context of redistricting, lets imagine each districting plan representing a \textit{state} and a random walk is being performed on this state space. As the walker traverses from one state to another, we collect each state. On terminating the walk, this collection constitutes the representative sample of the plans. Performing the \flip-based walk involves changing the assignment of individual geographic units along district borders as shown in \cref{ch4:fig:flip}. In the standard MCMC paradigm, altering this basic step adjusts the stationary distribution. \cref{ch4:fig:states} gives a rough approximation of the idea.

The remainder of this article is organized as follows. \cref{ch5:sec:background} shows how the theory of Markov Chain can be applied to the school redistricting problem in a graph-based setting. Next, \cref{ch5:sec:sampling} outlines how we can customize the \flip-based walks to incorporate the problem-specific constraints, objectives and criteria, thereby leading to three models. \cref{ch5:sec:experiment} details the experimental investigations performed in context of the real-world datasets of two school districts in the US. More specifically, we observe how these newly proposed models compare with the peer algorithms defined earlier.
Lastly, \cref{ch5:sec:discussions} concludes the article and provides some key takeaways and directions for future work.

\section{Background}\label{ch5:sec:background}

\subsection{Markov Chains}\label{ch5:sec:markovchain}
A \textit{Markov chain} or \textit{Markov process} is a stochastic model for moving between positions in a \textit{state space} as per a probabilistic transition rule under which the probability of arriving at a particular position at time $t+1$ depends only on the position at time $t$~\cite{gagniuc2017markov}. It is akin to a memoryless random walk. One classical example of a Markov chain is a simple random walk on graph: given a node $u$ at time $t$, the process selects a neighboring node of $u$ uniformly randomly at time $t+1$. In a more general setting, random walk can be performed on a weighted graph where each incident edge is associated with a different probability (of transitioning). Hence, any Markov chain can be accurately modeled as a (not necessarily simple) random walk on a (possibly directed) graph. Markov chains find use in plethora of applications, including speech recognition, network science, epidemiology and so on. In fact, they are the basis for general stochastic simulation methods like MCMC, which are used for simulating sampling from complex probability distributions~\cite{mcmc}.

Markov chains provide theoretical guarantees on convergence: there exists a unique stationary distribution for any ergodic Markov chain and any initial probability distribution can converge to that steady state if the transitions are iterated for sufficient number of steps. The \textit{mixing time} is the number of steps that it takes to be within an acceptable threshold of the steady state. %In most real-world applications, it is very uncommon to provide rigorous bounds on mixing time; instead, practitioners resort to a set of heuristic convergence tests and diagnostics.
% plagiarized
Here, we make an attempt to use Markov chains for a global exploration of the universe of valid redistricting plans. 
Stronger sampling and convergence theorems are available for reversible Markov chains. %, those for which the steady-state probability of being at state $u$ and transitioning to $v$ equals the probability of being at $v$ and transitioning to $u$ for all pairs $(u,v)$ from the state space.
\rewrite{
 In particular, Chikina et al. \cite{chikina2017assessing,chikina2020separating} presents a series of theorems showing that samples from reversible Markov chains admit conclusions about their likelihood of having been drawn from a stationary distribution $\pi$ long before the sampling distribution approaches $\pi$. In context of redistricting, this theory enables what we might call \textit{local search}~\cite{recombination}. As such, we compare the performance of the MCMC-based redistricting techniques that employ local search \cite{biswas2019regal,biswas2020spatial}.
}

\subsection{Redistricting as a graph partitioning problem}\label{ch5:sec:redistricting}
Given a geographic area, the first step is to construct a \textit{dual graph} or \textit{contiguity graph} $\graph = \bigl( \vertexset, \edgeset \bigr)$ with non-negative edge weights, $\omega: \edgeset \rightarrow \mathbb{R}\u{>0}$. %\rewrite{as shown in Figure XXX}.
Given the graph $\graph$, a districting plan seeks a partition $\mathbf{\Pi}$ of $\graph$ with blocks of nodes $\mathbf{\Pi} = \bigl(V\u{1},\ldots,V\u{K}\bigr)$  that induce connected subgraphs of $\graph$, i.e.,
$V\u{1} \cup \ldots \cup V\u{K} = \vertexset$, 
$V\u{i}$ is connected $\forall i \in \left\{1,2,\ldots,K\right\}$
and
$V\u{i} \cap V\u{j} = \phi \; \forall i,j \in \left\{1,2,\ldots,K\right\}, i \neq j$. 
In context of graph partitioning problem (GPP)~\cite{graphpartitioning}, one may imagine the output of a GPP is represented by a districting plan on $\graph$ described by an assignment function $\xi : \vertexset \rightarrow \left\{1,2,\ldots,K\right\}$, where $\xi\bigl(v\bigr)=i$ implies that node $v$ is assigned to subgraph $V \u{i}$. 
\rewrite{The nodes of $\graph$ that encode the spatial units may contain attributes like population, assignment to subgraph, and so on. The node attributes are useful in determining the validity of a plan. Similarly, the edges may also be associated with attributes like shared perimeter, etc.}

\subsection{Markov chains and Redistricting}
To be able to run Markov chains for the purpose of redistricting, we require a valid \textit{initial state} or districting plan and a rule or proposal method for transitioning from one state to another. It is customary to start the chain from the currently enacted plan. However, it is better to have multiple starting plans in order to create a more robust sample of plans that is independent of the starting plan. Hence, a rule of thumb is to start from a plan that is contiguous and have tolerably balanced partitions. In fact, one can imagine the redistricting as the problem of ``of sampling
from the state space of balanced partitions of a graph into a fixed number of connected
subgraphs''~\cite{connectedsampling}.
While this may appear to be a standard setting of Markov chain methods, there are some typical features of the redistricting problem that make it increasingly difficult for classical techniques to operate. 

\paragraph{Non-uniform sampling:} In a canonical Markov chain setting, it is desirable to sample from the uniform distribution of all graph partitions. However, the notion of uniform sampling does not fit the objective of the redistricting problem. As such, sampling from a nonunifrom distribution is preferred for a domain-specific reason: an overwhelming majority of plans in an ensemble drawn from an uniform distribution will be composed of arbitrarily-shaped districts (c.f. Figure 1 in \cite{recombination}). Such districts would be prohibitively non-compact and deemed useless for the purpose of redistricting. 
As a result, recent works on MCMC-based redistricting have focused on sampling from a nonuniform distribution~\cite{chikina2017assessing,pegden2017pennsylvania}.

\paragraph{Constrained problem structure:} The redistricting problem is further complicated by the presence of constraints like \textit{contiguity} and \textit{balance}. In fact, in problems like school redistricting, further challenge is added by the presence of fixed center in every district. These centers cannot be swapped between adjacent districts thereby highly restricting the transitions from one valid state to another.

In view of the aforementioned observations, we lay down the operational rules of developing a sampling technique in the view of the school redistricting problem.

\section{Sampling-based models}\label{ch5:sec:sampling}
A geographic region can be encoded by a contiguity graph $\graph = \bigl( \vertexset, \edgeset \bigr)$ as outlined in \cref{ch5:sec:redistricting}. A `districting plan' is a $K-$partition of $\graph$, i.e., a set of disjoint subsets of nodes. Let us denote the districting plan by $\partition = \left\{V \u 1, V \u 2, \ldots, V \u K\right\}$, where $V \u i$, correspond to a district. Henceforth, we shall use the terms plan and partition interchangeably. For the sake of convenience, we abuse the notation by assuming that $\partition$ also indicates an assignment function $\partition: \vertexset \rightarrow \left\{1,2,\ldots,K\right\}$, where $\partition(u) = i$ implies that $u \in V \u i$ for the plan $\partition$. For further convenience, we may overload the notation and write $\partition (u) = V \u i$.

A node $v$ is a neighbor of node $u$ if there is an edge $\bigl( u, v \bigr) \in \edgeset$. A node $u$ is called \textit{boundary node} if it has a neighboring node $v$ such that $\partition(u) \neq \partition(v)$. The set of boundary nodes is represented by $\delta {\u \vertexset} \partition = \left\{u: (u,v) \in \edgeset \textsf{ and } \partition(u) \neq \partition(v) \right\}$.  Correspondingly, an edge that connects two boundary nodes is called \textit{cut edge} and $\delta {\u \edgeset} \partition = \left\{(u, v) \in \edgeset: \partition(u) \neq \partition(v) \right\}$ is the set of cut edges. The contiguity graphs obtained from real-world datasets like school districts, usually have node-based features like count of residing students, demographic data, school assignment, and so on. We shall revisit them later on while discussing problem objectives and constraints.

Having defined the graph-based notations, we lay out the rule for transitioning between the different states. Let $\partition {\u K} (\graph)$ denote the set of all possible $K-$partitions on $\graph$.
The proposal distribution can be captured by a matrix $\mathbb{P}\u{\partition {\u k}} \in {[0,1]}^{|\partition \u{K}| \times |\partition \u{K}|}$. For simplicity, let us refer to this matrix as $\mathbb{P}$. The rows and columns of $\mathbb{P}$ are indexed by the states in $\partition {\u K}$. For instance, $\mathbb{P} {\u \stateR}$ indicate a row of $\mathbb{P}$ corresponding to state $\stateR$. Formally, $\mathbb{P} {\u \stateR}$ is ${[0,1]}^{|\partition \u K|}$-valued random variable where the values sum to one thereby indicating the probability of transitioning from state $\stateR$ to all other states. Likewise, the ${\mathbb{P}} {\u \stateR} (\stateS)$ is the probability of transitioning directly from state $\stateR$ to $\stateS$. Hence, a Markov process is instantiated, i.e., every successive state is drawn according to $\mathbb{P} {\u \stateR}$ given the present state is $\stateR$. From an application standpoint, both the states $\stateR$ and $\stateS$ correspond to a districting plan.

Note that the number of possible $K-$partitions of a graph $\graph$, i.e., $|\partition \u k|$ is an enormously large but finite number. Thus, computing the matrix $\mathbb{P}$ is practically impossible. One may imagine $\mathbb{P}$ to be a stochastic matrix and thus the proposal distribution as a stochastic algorithm for modifying the assignment of nodes. This helps to circumvent the computation of  $\mathbb{P}$. Building on this perspective, we discuss the flip-based proposal for transitioning between the different states. We closely follow \cite{recombination} for outlining the next steps.

\subsection{\flip-based proposal}\label{ch5:sec:flip}\label{ch5:sec:model}
\begin{figure}
    \begin{center}
        \includegraphics[keepaspectratio, width=0.75\textwidth]{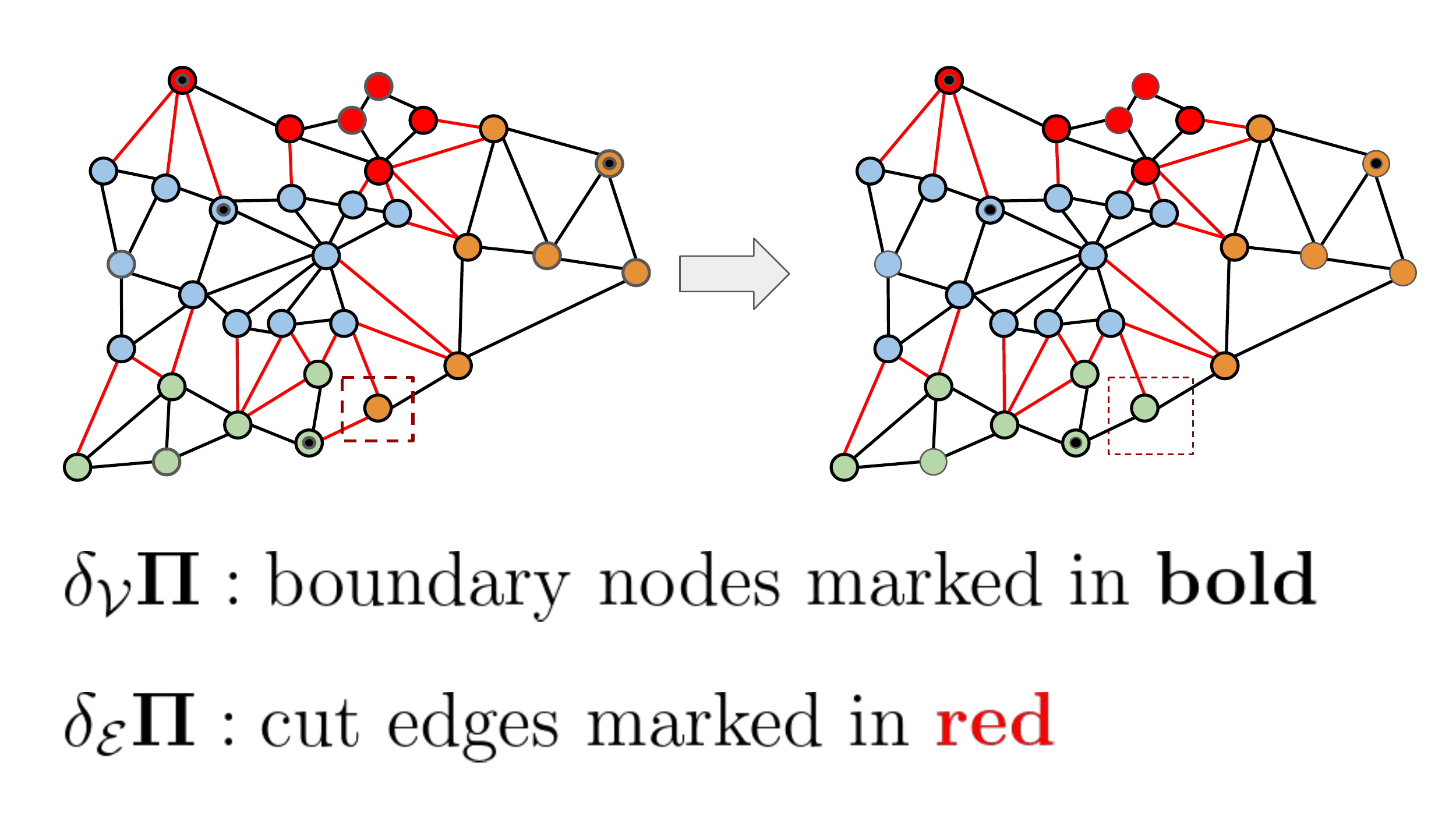}
    \caption{The flip based proposal samples uniformly on the space of boundary nodes and cut edges.}
    \label{ch4:fig:boundary}
    \end{center}
\end{figure}

The \flip~proposal is the most basic type of local search where the assignment of a node is altered while maintaining the contiguity of the plan. This is identical to the local search technique employed by techniques like \regal~and illustrated in \cref{ch4:fig:boundary}. We will build on these up next.

The first step is to select a node for altering its assignment. In order to maintain the contiguity of the plan, it makes sense to select a node from $\delta {\u \vertexset} \partition$ or an edge from $\delta {\u \edgeset} \partition$. However, due to their varying degrees it might introduce non-uniformity in selecting the node. To circumvent this problem, the set of $(u, V \u i)$ pairs is sampled uniformly, where $u \in \delta {\u \vertexset} \partition$ and there exists a cut edge $(u, v) \in \delta {\u \edgeset} \partition$. This helps to uniformly select among the partitions differing only by the assignment of a single boundary node~\cite{recombination}. The transition probability of the corresponding Markov chain is given as follows

\footnotesize
\begin{equation}
    \mathbb{P}{\u \stateR}{( \stateS )} =
    \left\{\begin{matrix}
\frac{1}{|\left\{(v,\stateR(w)):(v,w)\in \delta {\u \edgeset} \stateR)\right\}|} & \text{if }
|\{\stateR(u) \neq \stateS(u): u \in \delta {\u \vertexset} \stateR\}| = 1
\textsf{ and }
|\{ \stateR(u) \neq \stateS(u): u \in \delta {\u \vertexset} \stateR \}| = 0 \\ 
0 & \textsf{otherwise.}\qquad \qquad \;
\end{matrix}\right.
\end{equation}
\normalsize
This is akin to performing a simple random walk on $\partition {\u k} (\graph)$ where two partitions (corresponding to states $\stateR$ and  $\stateS$) differing by a single boundary node are connected. Thus, the Markov Chain is reversible, which lends itself to theoretical guarantees outlined in~\cite{chikina2017assessing}. \rewrite{Note that the stationary distribution of this Markov chain is non-stationary since the each plan is weighted proportionally to the number of (node, district) pairs in its boundary.}

Once the node has been selected, a canonical flip operation would result in shrinking the size of one district (say, \textit{donor} by 1 while adding to the size of another district (say, \textit{recipient)}. This would result in the \textit{recipient} district maintaining its contiguity but may break the contiguity in the \textit{donor} district. To avoid this, a rejection sampling-based approach is adopted, i.e., only flip proposals that maintain its contiguity in both the \textit{donor} and \textit{recipient} district are accepted.  This completes the second step.

Interestingly, adoption of rejection sampling does not violate the reversibility of the Markov chain. Only the random walk is confined to a restricted state space. From a practical standpoint, rejection sampling since it is far more computationally efficient to determine if a particular plan can be accepted than to enumerate the entire set of adjacent plans at every step of the random walk. If a plan fails contiguity the contiguity check, a new plan is proposed from the previous state. This step is repeated till a plan finally passes the check. This flexibility of the rejection sampling enables us to incorporate additional constraints, besides contiguity, as discussed next. 

\subsection{Incorporating preferences}\label{ch5:sec:constraints}
In a redistricting scenario, the preferences and priorities of practitioners are enacted through a series of modeling decisions. \rewrite{In view of the school redistricting problem, we operationalize a set of criteria next.}

\paragraph{Contiguity:} The contiguity-based rejection sampling is instantiated through the $\constraint \u 0$ function, which evaluates to $True$ is the newly proposed plan is contiguous, otherwise it returns $False$.

\paragraph{No vanishing districts:} For a partition $\partition$ to be valid, all the districts should have a finite size. This is enacted by the following constraint function
\begin{equation}
    \constraint {\u 1} \left( \partition \right)
    =\left\{\begin{matrix}
    True & |V_i|>0\; \forall V_i\in \partition,\;i=1,2,\ldots,K\\ 
    False & \textsf{otherwise                     }
    \end{matrix}\right. .
    \label{ch5:eq:constraint1}
\end{equation}

\paragraph{Single center districts:} Every district should contain only one center node. In school redistricting, the center nodes correspond to the schools. It can be modeled as
\begin{equation}
    \constraint {\u 2} \left( \partition \right)
    =\left\{\begin{matrix}
    True & C(V\u i)=1\; \forall V_i\in \partition,\;i=1,2,\ldots,K\\ 
    False & \textsf{otherwise                     }.
    \end{matrix}\right.,
    \label{ch5:eq:constraint2}
\end{equation}
where $C(V\u i)$ counts the number of center nodes in district $i$.

\paragraph{Compact districts:} Many school districts have a \textit{compactness} criteria indicating their preference for regular-shaped districts. Here, compactness is used as a proxy for reduced commute distance or travel time. The Polsby-Popper metric, a popular isoperimetric score for measuring compactness, is used here. The compactness of districts is ensured through a $L^{-1}$ aggregation of the compactness scores of the districts as defined next.
\begin{equation}
    \constraint {\u 3} \left( \partition \u t \right)
    =\left\{\begin{matrix}
    True & L^{-1} _{\u {\textsf{PP}}} \left( \partition \u t \right) \geq L^{-1} _{\u {\textsf{PP}}} \left( \partition \u{t-1} \right) + \varepsilon\\ 
    False & \textsf{otherwise                     }
    \end{matrix}\right.,
    \label{ch5:eq:constraint3}
\end{equation}
where $\partition \u t$ is the partition at step $t$, $\varepsilon$ is a thresholding parameter set to 0.05 and $L^{-1} _{\u {\textsf{PP}}}$ is the harmonic mean of the Polsby-Popper scores of the individual districts calculated as 
\begin{equation}
    L^{-1} _{\u {\textsf{PP}}} \left( \partition \right) = \frac{K}{\frac{1}{PP(V \u 1)}+ \frac{1}{PP(V \u 2)} + \cdots + \frac{1}{PP(V \u K)}}.
    \label{ch5:eq:harmonic}
\end{equation}
$PP(V\u i)$, $i=1,2,\ldots, K$, captures the Polsby-Popper compactness score of the districts.

This constraint function acts like a self-configuring lower bound that heavily penalizes the most non-compact district in a partition~\cite{pegden2017pennsylvania}.

\paragraph{Balanced districts:} The presence of overpopulated or under-enrolled schools is one of the main reasons why school boundaries are frequently redrawn. As such, it is highly desirable that the attending population in the schools are balanced with respect to their program capacity. This is implemented through the constraint function 
\begin{equation}
    \constraint {\u 4} \left( \partition \u t \right)
    =\left\{\begin{matrix}
    True & {Imb} \left( \partition \u t \right) \leq {Imb} \left( \partition \u 0 \right) \\ 
    False & \textsf{otherwise                     }
    \end{matrix}\right.,
    \label{ch5:eq:constraint4}
\end{equation}
where $\partition \u t$ is the partition at step $t$,  $\partition \u 0$ is the starting partition and ${Imb}()$ function measures the degree of imbalance in the districts of the partition as 
\begin{equation}
    Imb \left( \partition \right) =
        \sum\limits_{i=1}^{K}{
            \left | 1 - 
            \frac{
            \sum\limits_{u \in V \u i} \textsf{Pop}\left( u \right)
            }{
            \sum\limits_{u \in V \u i} \textsf{Cap}\left( u \right)
            } 
            \right |
            }.
    \label{ch5:eq:imbalance}
\end{equation}
where $\textsf{Pop}\left( u \right)$ and $\textsf{Cap}\left( u \right)$ captures the count of  the residing student population and the school capacity of node $u$. The student population is computed as three school levels$-$elementary, middle and high. Note that for majority of the nodes (except center nodes) the value of $\textsf{Cap}(u)$ is 0 since they don't contain a school.

\paragraph{Well-balanced and compact districts:} Ideally, we would like the districts to be well-balanced with compact boundaries as a culmination of the two scenarios discussed earlier.
\begin{equation}
    \constraint {\u 5} \left( \partition \u t \right)
    =\left\{\begin{matrix}
    True & {J} \left( \partition \u t \right) \leq {J} \left( \partition \u 0 \right) \\ 
    False & \textsf{otherwise                     }
    \end{matrix}\right.,
    \label{ch5:eq:constraint5}
\end{equation}
where ${J}()$ is a dispersion function that gives a weighted score to a partition by weighing in the degree of imbalance and non-compactness as 
\begin{equation}
    J\bigl( \partition \bigr)
            =
            \lambda
            \sum\limits_{i=1}^{K}{
            \left | 1 - 
            \frac{
            \sum\limits_{u \in V \u i} \textsf{Pop}\left( u \right)
            }{
            \sum\limits_{u \in V \u i} \textsf{Cap}\left( u \right)
            } 
            \right |
            }
            \; + \;
            (1-\lambda)
            \sum\limits_{i=1}^{K}{
            \left |
            1 - 
            \frac{ 4 \pi \cdot
            \textsf{Area} \bigl( \bigcup\limits_{u \in V \u i} u \bigr)
            }{
            \textsf{Peri}\left[ \bigl( \bigcup\limits_{u \in V \u i} u \bigr)\right] ^2
            }
            \right |
                }
    \label{ch5:eq:dispersion}
\end{equation}
$\constraint \u 5$ acts like a randomized greedy search that samples plans of higher quality as compared to the starting plan. 

Depending on the preference of the modeler, any subset of these aforementioned constraints can be selected to create a consolidated constraint function
\begin{equation}
    \constraint \left( \partition \right)=
    \left\{\begin{matrix}
        True & \constraint {\u i} \text{ is }{True}\;\forall\;i \subseteq \{1,2,\ldots,C\} \\ 
        False & \textsf{otherwise}\qquad\qquad\qquad\;
    \end{matrix}\right.,
    \label{ch5:eq:constraint_function}
\end{equation}
where $C$ is the number of constraint functions under consideration and $i$ is any subset of those functions.
A plan $\partition$ is considered admissible if it passes the constraint check, i.e., $\constraint(\partition)$ evaluates to $True$. As mentioned earlier, this is akin to a random walk in a restricted state space. 

It is at the discretion of the modeler whether to accept a plan as it is if it passes the constraint check or to include further restrictions. This gives rise to two acceptance conditions:
\begin{itemize}
\item \textbf{Always accept:} As the name suggests, the newly proposed plan is always accepted irrespective of its quality.
\item \textbf{Accept improving:} This is a greedy acceptance condition which only accepts a new plan if it is better than the present plan, i.e., $ {J} \left( \partition \u t \right) < {J} \left( \partition \u {t-1} \right)$. It is equivalent to the greedy selection mechanism applied in techniques like \regal~\cite{biswas2019regal} and~\spatial~\cite{biswas2020spatial}.
\end{itemize}

\subsection{Models}\label{ch5:sec:models}
In this section, we make use of the constraint functions and the acceptance conditions to define three sampling models that sample new plans by simulating the $\flip$ proposal. Note that there are more sophisticated proposals like Recombination proposed for the political districting problem~\cite{recombination}. However, our simulations on real-world datasets of school districts have revealed that Recombination does poorly at satisfying the $\constraint \u 0$, especially when the center nodes are adjacent to each other.  Hence, we adopt the the simple $\flip-$based transitions.

We ensure that the newly sampled plans have compact, contiguous, non-vanishing and single-center districts by enforcing the constraint functions $\constraint {\u 0} - \constraint {\u 3}$. These models only differ in the constraint functions $\constraint \u 4$ and $\constraint \u 5$, and the  acceptance conditions. They are named as follows:
\begin{itemize}
    \item \textsf{B}alanced, \textsf{A}lways \textsf{A}ccept (\modelA)
    \item \textsf{B}alanced and \textsf{C}ompact, \textsf{A}ccept (\modelB)
    \item \textsf{A}ccept \textsf{I}mproving \textsf{O}bjective (\modelC)
\end{itemize}
\cref{ch5:tab:models} outlines the models and their configurations succinctly. We kept the number of $\flip$ proposals to 10 million for getting a better representation. These sampling models serve two purposes:
\begin{itemize}
    \item They serve as a gold standard for local search techniques like \regal~and \spatial. This is based on the findings by~\cite{chikina2017assessing}.
    \item They provide diagnostics tools for drawing useful conclusions about the intermediate states while arriving at a final districting plan. 
\end{itemize}

\begin{table}
    \centering
    \caption{The configuration of the proposed sampling-based models for generating districting plans.}
    \small
    \begin{tabular}{@{}c|c|c|c@{}}
    % \hline
    \toprule
     %\hline
    %\begin{tabular}{|c||c|c|c||c|c|c||c|c|c|}
    %\hline
    \multirow{2}{*}{
    \diagbox{Configuration}{Models}
    }  & 
    \multirow{2}{*}{\modelA} &
    \multirow{2}{*}{\modelB} &
    \multirow{2}{*}{\modelC}
    \\ 
    %\cline{2-4}
    &   &    &    \\ 
    %\hline 
    \midrule
    \textsf{Contiguity} $(\constraint \u 0)$ & \checkmark & \checkmark & \checkmark
    \\ %\hline
    \textsf{Non-vanishing} $(\constraint \u 1)$ &
    \checkmark & 
    \checkmark & 
    \checkmark 
    \\ % \hline
    \textsf{Single-center} $(\constraint \u 2)$ &
    \checkmark &  
    \checkmark &  
    \checkmark 
    \\ % \hline
    \textsf{Compact} $(\constraint \u 3)$ &
    \checkmark & 
    \checkmark & 
    \checkmark
    \\ % \hline
    \textsf{Balanced} $(\constraint \u 4)$ &
     \checkmark
     & 
     &
    
    \\ % \hline
    \textsf{Balanced and compact} $(\constraint \u 5)$ &
     &
     \checkmark
     &
    \\ % \hline
    \midrule
    \textsf{Always accept} &
    \checkmark & \checkmark &
    \\ 
    \textsf{Accept improving}
    & & & \checkmark
    \\  \bottomrule
    \end{tabular}
    %\par}
    \normalsize
    \label{ch5:tab:models}
\end{table}

\section{Experimentation}\label{ch5:sec:experiment}
The three sampling-based methods were built on top of the \textsf{GerryChain} library\footnote{Available at \url{https://github.com/mggg/GerryChain}.} developed and maintained by the Metric Geometry and Gerrymandering Group\footnote{\url{https://www.mggg.org/}} (MGGG). Some of the implementations are based on personal communication with the MGGG team. The code is available at \url{https://github.com/subhodipbiswas/SamplingbasedSchoolRedistricting}. 
Next, we empirically investigate the proposed sampling-based methods on the school redistricting problem. 
To this effect, the dataset details are provided in \cref{ch5:sec:dataset} followed by the performance metrics in 
\cref{ch5:sec:evaluation}. \cref{ch5:sec:empirical} details the empirical investigations and the results are discussed in \cref{ch5:sec:results}.

\subsection{Dataset}
\label{ch5:sec:dataset}
The study was performed on two school districts (counties) located in the mid-Atlantic region of the USA. These school districts have seen recent population growth in certain areas, thereby making the problem challenging for the redistricting algorithms. The following GIS data attributes of both the districts were used for experimentation.
\begin{itemize}
    \item \textbf{SPAs}: The location coordinates of the spatial units along with aggregated student count at different school levels (Elementary, Middle and High).
    \item \textbf{Schools}: The location of the school building, school level, and its program capacity.
\end{itemize}
In comparison to the earlier works~\cite{biswas2019regal,biswas2020spatial}, we used the data for the  school year $2020-2021$. We resorted to using the new dataset since many parents unenrolled their children from the public schools at the onset of the COVID-19 pandemic resulting in more population imbalance. This would present more challenging problem scenarios for the redistricting algorithms. Table~\ref{ch5:tab:data} presents the summary statistics of this new dataset.
  
\begin{table}[htp!]
\centering
  \caption{Summary statistics of the school districts for the school year 2020-21.}
  \label{ch5:tab:data}
  \rewrite{
  \begin{tabular}{c|c|ccc}
    \toprule
    \multirow{2}{*}{District} & \multirow{2}{*}{\#SPAs ($N$)}  & \multicolumn{3}{c}{\#Schools ($K$)}
    \\ \cline{3-5}
    & & Elementary  & Middle & High
    \\ \hline
    \lcps &  453 &  57 & 16 & 16
    \\ 
    \fcps & 1313 & 138 & 26 & 24\\ 
    \bottomrule
  \end{tabular}
}
\end{table}

\subsection{Evaluation metrics}
\label{ch5:sec:evaluation}
The solution to the school redistricting problem is a districting plan or a zoning configuration of the school boundaries. For evaluating the quality of the solutions, we utilized two performance metrics$-$\textit{balance} and \textit{compactness} defined next. 
    Compact school boundaries often translate to proximal schools that students can walk to and thereby lower the transportation cost incurred by the school district.

\begin{itemize}
    \item \textbf{Balance} measures the average balance between a school's program capacity and the number of students residing within its boundary. It is calculated as
    \begin{equation}
    \text{bal}\left(\partition\right) = 
    100 \times
            \left | 1 - 
            \sum\limits_{i=1}^{K}{
            \left | 1 - 
            \frac{
            \sum\limits_{u \in V \u i} \textsf{Pop}\left( u \right)
            }{
            \sum\limits_{u \in V \u i} \textsf{Cap}\left( u \right)
            } 
            \right |
            }
            \right |
    \end{equation} 
    We penalized both under-enrolled and overburdened schools equally with respect to the capacity of schools. This is an important metric for school planners since most of the boundary changes occur to achieve a better balance in schools.
    
    \item \textbf{Compactness} measures how tightly a school's boundary is packed on an average with respect to its perimeter. A scaled Polsby-Popper score~\cite{polsby1991third} is used to measure compactness as
    \begin{equation}
    \text{com}\left(\partition\right) = 
     \frac{100}{K}
        \sum\limits_{i=1}^{K}{
            \left |
            1 - 
            \frac{ 4 \pi \cdot
            \textsf{Area} \bigl( \bigcup\limits_{u \in V \u i} u \bigr)
            }{
            \textsf{Peri}\left[ \bigl( \bigcup\limits_{u \in V \u i} u \bigr)\right] ^2
            }
            \right |
        }
    \end{equation} 

\end{itemize}

Both these metrics can be interpreted as percentage scores that lie in the range of $\left[0, 100\right]$. The higher the value, the better is the quality of the plan.

\subsection{Empirical investigations}\label{ch5:sec:empirical}
With the help of the sampling-based techniques, one may seek answers to the following research questions:
\begin{itemize}
    \item[\textbf{Q1}.] How does these sampling-based techniques compare to the local search methods defined earlier?
    
    \item[\textbf{Q2}.] Do the sampling-based techniques provide any useful diagnostics on the school redistricting problem?
    
    %\item[\textbf{Q3}.] \rewrite{Does the initial state affect the final state in the sampling-based technique?}
 
\end{itemize}
    
For answering these questions, we simulate different redistricting model on 6 problem instances$-$ 2 datasets (Districts \lcps~and \fcps) each containing 3 school levels (elementary, middle and high). For every problem instance, 25 trial runs are simulated. Next, we focus on the experimental settings in relation to the above questions.

\begin{itemize}
    \item[\textbf{E1}.] To answer \textbf{Q1}, we instantiate an ensemble of 25 trial plans for each problem instance such that the trial plans mirror the present districting plans. We noticed that in some of the problem instances, the present districting plan is lacking in contiguity. As such we had to fix those plans using the repair mechanism outlined in \cite{biswas2020spatial}. This may result in the plans for each trial run differing slightly from one another.
    For comparative study, we including the following redistricting models$-$
    the three sampling-based methods, \spatial~\cite{biswas2020spatial} and the \regal-based local search techniques \cite{biswas2019regal}, i.e., Stochastic Hill Climbing (\shc), Tabu Search (\ts) and Simulated Annealing (\sa). 
    We ensure fair comparison by making every redistrict model start with the same trial solution. Only for \spatial, which is a population-based model, we randomly pick 10 plans from the ensemble.
    
    \item[\textbf{E2}.] To answer \textbf{Q2}, we simulate the sampling-based methods on each problem instance for three types of initialization schemes$-$random, distance-based and present. While simulating trial runs, we run diagnostics to keep a track of few problem-specific metrics. We analyze the run-time diagnostics to see if they throw light on useful problem-specific properties that may help in the design of more efficient search techniques later on.
    
\end{itemize}

For all the peer algorithms, the model parameters are set to default unless otherwise stated.

\subsection{Results}\label{ch5:sec:results}
\paragraph{E1:} The results are summarized in \cref{ch5:tab:comparison}. The redistricting models are grouped by their algorithm class. On District \lcps, \ts~has an edge over the class of local search-based methods. However, the memetic algorithm \regal~is a more balanced one since it achieves balance scores comparable to \ts~and does better in terms of compactness. This is expected since \spatial~combines the benefits of local search with a spatial recombination operator. Across both the districts, \ts~and~\spatial~are the top-performing models.

For evaluating the results of the sampling-based methods, we update the compactness and balance scores when a greedy criterion, i.e., ${J} \left( \partition \u t \right) \leq {J} \left( \partition \u {t-1} \right)$ is satisfied. Note that this greedy criterion is just implemented for bookkeeping and does not affect the acceptance condition of the sampling-based models. From the results, it is evident that the explorative variants, namely \modelA~and \modelB, do not generate good plans. Their performance is especially poor for District \fcps, which is a much larger-sized problem instance than District \lcps. On the other hand, \modelC, aided by the greedy acceptance condition is significantly better-performing than its other sampling counterparts, especially in terms of compactness. In fact, \modelC~resembles a local search like \shc.
It is important to note that the purpose of models \modelA~and~\modelB~is to perform exploration in the state space of plans and help in running diagnostics. \modelC~on the other hand was created with the intention to create balanced and compact districts. They serve their purpose well enough. In the next subsection, we will see how the explorative models help in infering important characteristics about the problem.

\begin{table}
    \centering
    \caption{Performance of the baseline models in designing school boundaries of both the districts.}
    \label{ch5:tab:comparison}
    \setlength{\tabcolsep}{1pt}
    \footnotesize
    %{\fontsize{9.0pt}{10.0pt}\selectfont
    % \begin{tabular}{|c|c|c|c|c|c|c|c|c|c|}
    \begin{tabular}{@{}c|cc|cc|cc@{}}
    % \hline
    \toprule
    \multicolumn{7}{c}{District \lcps}
    \\ 
    \hline
    \multirow{2}{*}{
    Models
    }  & 
    \multicolumn{2}{c}{Elementary School} &
    \multicolumn{2}{c}{Middle School} &
    \multicolumn{2}{c}{High School}
    \\ \cline{2-7} &
        Balance     &    Compactness    &
        Balance     &    Compactness    &
        Balance     &    Compactness    \\ 
        %\hline 
        \midrule
    \texttt{Existing} &
    83.5020 $\pm$ 0.0000 & 32.5344 $\pm$ 0.0000 &  % ES
    89.7379 $\pm$ 0.0000 & 26.7671 $\pm$ 0.0000 &  % MS
    87.0786 $\pm$ 0.0000 & 27.3452 $\pm$ 0.0000  % HS
    \\
        \midrule
    SHC &
    88.0565 $\pm$ 0.3398 & 40.4219 $\pm$ 0.6413 &  % ES
    92.6503 $\pm$ 0.1009 & 37.9347 $\pm$ 1.1750 &  % MS
    97.5288 $\pm$ 0.6344 & 33.3857 $\pm$ 1.6935    % HS  
    \\ % \hline
    SA &
    87.9809 $\pm$ 0.5044 & 40.9826 $\pm$ 1.0907 &  % ES
    92.5345 $\pm$ 0.2965 & 37.1108 $\pm$ 2.1572 &  % MS
    97.3246 $\pm$ 0.5107 & 33.0759 $\pm$ 1.8799    % HS  
    \\ % \hline
    TS &
    88.2290 $\pm$ 0.3177 & 40.4481 $\pm$ 0.5532 &  % ES
    \textbf{92.7145} $\pm$ 0.0019 & 38.1273 $\pm$ 0.1339 & % MS
    97.3869 $\pm$ 0.1536 & 32.7663 $\pm$ 0.3416 % HS 
    \\ % \hline
    \midrule     
    \modelA &
    83.0738 $\pm$ 0.6002 & 32.4948 $\pm$ 1.0189 &  % ES
    92.1247 $\pm$ 0.3935 & 28.4179 $\pm$ 0.8139 &  % MS
    94.5031 $\pm$ 0.7808 & 26.3284 $\pm$ 1.3810    % HS  
    \\          
    \modelB &
    84.1403 $\pm$ 0.5138 & 32.2777 $\pm$ 1.0458 &  % ES
    92.2637 $\pm$ 0.3897 & 29.6189 $\pm$ 1.0030 &  % MS
    95.5958 $\pm$ 0.8448 & 26.2145 $\pm$ 1.6023    % HS  
    \\ % \hline
    \modelC &
    87.7953 $\pm$ 0.4931 & 41.0278 $\pm$ 0.7990  &  % ES
    92.6865 $\pm$ 0.0594 & 37.1659 $\pm$ 0.8954  &  % MS
    96.6412 $\pm$ 1.0957 & 33.7013 $\pm$ 1.7627     % HS  
    \\ 
    \midrule    
    \spatial &
    \textbf{88.2474} $\pm$ 0.3008 & \textbf{41.6610} $\pm$ 0.9948  &  % ES
    92.6639 $\pm$ 0.0642 & \textbf{39.9138} $\pm$ 0.7408  &  % MS
    \textbf{98.0286} $\pm$ 0.2052 & \textbf{35.2453} $\pm$ 1.2159 % HS  
    \\
    \bottomrule
    \\
    \multicolumn{7}{c}{District \fcps} 
    \\ 
    \hline
    \multirow{2}{*}{
    Models
    }  & 
    \multicolumn{2}{c}{Elementary School} &
    \multicolumn{2}{c}{Middle School} &
    \multicolumn{2}{c}{High School}
    \\ \cline{2-7} &
        Balance     &    Compactness    &
        Balance     &    Compactness    &
        Balance     &    Compactness    \\ 
        %\hline 
        \midrule
    \texttt{Existing} &
    82.3835 $\pm$ 0.0000 & 35.9234 $\pm$ 0.0000 &  % ES 
    84.2310 $\pm$ 0.0000 & 27.7096 $\pm$ 0.0000 & % MS 
    86.9541 $\pm$ 0.0000 & 26.8006 $\pm$ 0.0000   % HS
    \\
        \midrule
    SHC &
    94.2043 $\pm$ 0.5056 & 35.3233 $\pm$ 0.6058 &  % ES
    90.7112 $\pm$ 1.3864 & 29.9091 $\pm$ 1.3810 &  % MS
    91.9250 $\pm$ 0.0950 & 32.1575 $\pm$ 1.0527    % HS  
    \\ 
    SA &
    95.1757 $\pm$ 0.5959 & 33.9204 $\pm$ 0.9149 &  % ES
    91.4492 $\pm$ 0.6952 & 28.2519 $\pm$ 1.6041 &  % MS
    92.2339 $\pm$ 0.4015 & 30.4304 $\pm$ 1.2035    % HS  
    \\ 
    TS &
    93.9882 $\pm$ 0.2143 & 37.4500 $\pm$ 0.4358 &  % ES
    89.8580 $\pm$ 0.6495 & \textbf{31.9823} $\pm$ 1.0187 &  % MS
    \textbf{92.0073} $\pm$ 0.1227 & 34.3255 $\pm$ 0.7516    % HS 
    \\ % \hline
    \midrule  
    \modelA &
    80.1400 $\pm$ 0.4253 & 32.3159 $\pm$ 0.4638 &  % ES
    81.5796 $\pm$ 1.0063 & 22.8711 $\pm$ 0.9850 &  % MS
    87.7556 $\pm$ 0.6411 & 22.6101 $\pm$ 0.7538    % HS  
    \\
    \modelB &
    81.3120 $\pm$ 0.3927 & 32.0952 $\pm$ 0.5303 &  % ES
    83.1391 $\pm$ 0.9270 & 22.4642 $\pm$ 0.7627 &  % MS
    88.9047 $\pm$ 0.5496 & 22.2048 $\pm$ 0.7405    % HS  
    \\ % \hline
    \modelC &
    94.3157 $\pm$ 0.5046 & \textbf{41.0278} $\pm$ 0.7990  &  % ES
    91.3974 $\pm$ 0.9010 & 31.1680 $\pm$ 1.5590  &  % MS
    91.8972 $\pm$ 0.1067 & 33.0105 $\pm$ 1.1274     % HS  
    \\ 
    \midrule  
    \spatial &
    \textbf{95.3731} $\pm$ 0.3617 & 36.1492 $\pm$ 0.7468  &  % ES
    \textbf{91.7854} $\pm$ 0.2029 & 31.8754 $\pm$ 1.3194  &  % MS
    91.9185 $\pm$ 0.1156 & \textbf{34.6683} $\pm$ 0.9838  % HS   
    \\
      \bottomrule
    \end{tabular}
    \normalsize
\end{table}

\begin{figure}
    \centering
    \begin{subfigure}[b]{\linewidth}
        \includegraphics[width=0.3\textwidth, keepaspectratio]{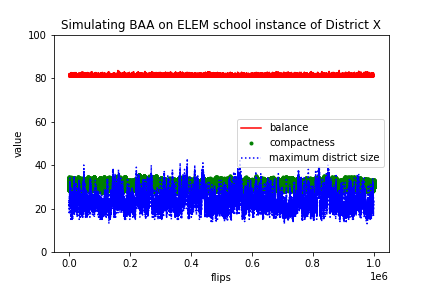}
    ~    
        \includegraphics[width=0.3\textwidth, keepaspectratio]{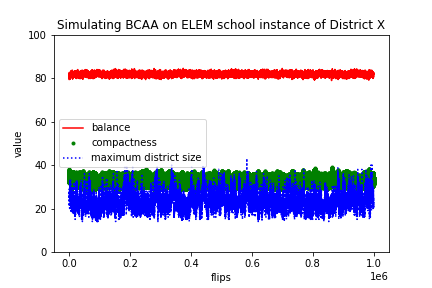}
    ~    
        \includegraphics[width=0.3\textwidth, keepaspectratio]{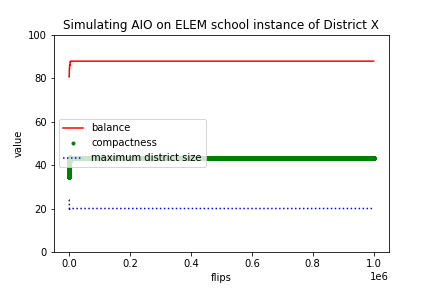}
        \caption{Elementary school}
    \end{subfigure}

    \begin{subfigure}[b]{\linewidth}
        \includegraphics[width=0.3\textwidth, keepaspectratio]{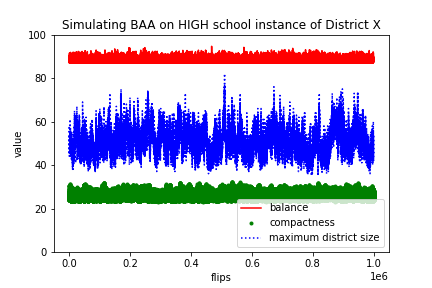}
    ~    
        \includegraphics[width=0.3\textwidth, keepaspectratio]{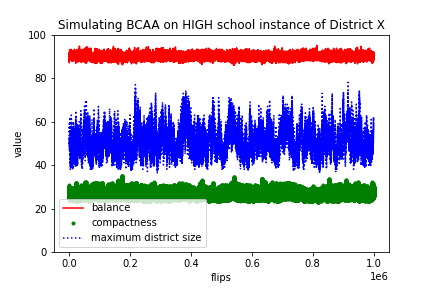}
    ~    
        \includegraphics[width=0.3\textwidth, keepaspectratio]{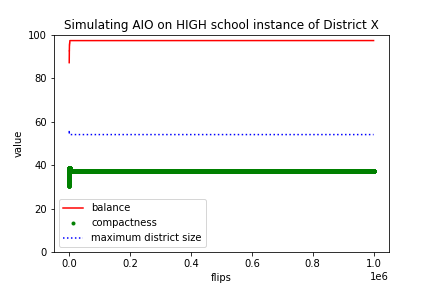}
        \caption{High school}
    \end{subfigure}
     \caption{MCMC simulation on District \lcps~on elementary and middle school instances. The explorative tendency of \modelA~and \modelB~help to uncover a wide range of values for metrics like maximum size of a district.}
\label{ch5:fig:operators}
\end{figure}

\paragraph{E2:} The result of MCMC diagnostics are shown in \cref{ch5:fig:operators}. We notice that explorative variants like \modelA~and \modelB~are actually useful in getting a range of metrics like maximum district size and so on. The greedy \modelC~is not suited for such simulations as it spans a very narrow range of values. We notice that the value of district size rarely crosses 40 and 80 for elementary and high school, respectively.

A similar diagnostics was run to keep a track of the co-occurence of nodes in the districts. To do this, we first encoded a binary representation of a plan $\partition$ in the form of a binary matrix $\assignment$, i.e., $\assignment \in \{0, 1\} ^{|V| \times |V|}$. Then we performed a simple book-keeping by sampling the partition $\partition \u t$ at every $1000^{th}$ flip as follows: if two nodes $(u,v)$ co-occur in a district of the partition, we mark $X \u {uv} = 1$ such that $\partition{u} < \partition{v}$. This updating enforces an ordering and helps to determine the number of non-zero entries after 10M flips have been performed. On assessing the number of non-zero entries in the binary co-occurence matrix, we notice that for District \lcps~,the matrix $\assignment$ is empty (filled with 0) at least 80\%, 65\% and 60\%, for elementary, middle and high school instances, respectively. Similarly, for District \fcps, at least 92\%, 73\% and 76\% of $\assignment$ is empty for elementary, middle and high school instances, respectively. This indicates the sparsity of the binary representation of a plan $\partition$.

These MCMC-based diagnostics serve as a rough estimation bounds for these metrics and can be used to prune down the combinatorial search space further than in a vanilla implementation.
More on this is discussed in Chapter 6 of \cite{biswas2022spatial}.

\section{Conclusion}\label{ch5:sec:discussions}
In this article, we show the utility of MCMC-based sampling techniques in view of the school redistricting problem. These techniques were built using the well-known \textsf{GerryChain} package. We further show the flexibility of these sampling-based techniques$-$they can be used for optimization as in the \modelC~model or for diagnostics as in models \modelA~and \modelB. The \modelC~model can serve as a gold standard for local search techniques developed earlier. The explorative models like \modelA~and \modelB~help us to determine a range of values of problem parameters. The parameters and the observations inferred from the MCMC diagnostics may be useful in devising exact methods for solving this problem.

\section*{Acknowledgments}
We are grateful to Stephen C Billups and Robert Hildebrand for helpful discussions. This research is supported in part by National Science Foundation (NSF) grants DGE-1545362 and IIS-1633363. \textbf{Disclaimer:} The views and conclusions contained herein are those of the authors and should not be interpreted as necessarily representing the official policies or endorsements, either expressed or implied, of any school board, NSF, or the U.S. Government.

\bibliographystyle{unsrt}  
\bibliography{references}

\end{document}